\documentclass{article}
\usepackage{graphicx}
\usepackage{amsmath}
\usepackage{caption}
\usepackage{subcaption}
\usepackage[nonatbib, preprint]{neurips_2022}
\usepackage{biblatex}
\newcommand{\mycite}[1]{ (\textit{\citeauthor{#1}\cite{#1})} }
\usepackage[utf8]{inputenc} % allow utf-8 input
\usepackage[T1]{fontenc}    % use 8-bit T1 fonts
\usepackage{hyperref}       % hyperlinks
\usepackage{url}            % simple URL typesetting
\usepackage{booktabs}       % professional-quality tables
\usepackage{amsfonts}       % blackboard math symbols
\usepackage{nicefrac}       % compact symbols for 1/2, etc.
\usepackage{microtype}      % microtypography
\usepackage{xcolor}         % colors/
\title{What Do Graph Convolutional Neural Networks Learn?}
\author{
    Sannat S Bhasin\\
    University of Toronto, St. George \\
    Toronto, Canada \\
    \texttt{sannat.bhasin@mail.utoronto.ca} \\
    \And
    Vaibhav Holani \\
    University of Toronto, St. George \\
    Toronto, Canada \\
    \texttt{vaibhav.holani@mail.utoronto.ca} \\
    \AND
    Divij Sanjanwala \\
    University of Toronto, St. George \\
    Toronto, Canada\\
    \texttt{divij.sanjanwala@mail.utoronto.ca} \\
}
\begin{document}
\maketitle
\begin{abstract}
Graph neural networks (GNNs) have gained traction over the past few years for their superior performance in numerous machine learning tasks. Graph Convolutional Neural Networks (GCN) are a common variant of GNNs that are known to have high performance in semi-supervised node classification (SSNC), and work well under the assumption of homophily. Recent literature has highlighted that GCNs can achieve strong performance on heterophilous graphs under certain "special conditions". These arguments motivate us to understand why, and how, GCNs learn to perform SSNC. We find a positive correlation between similarity of latent node embeddings of nodes within a class and the performance of a GCN. Our investigation on underlying graph structures of a dataset finds that a GCN's SSNC performance is significantly influenced by the consistency and uniqueness in neighborhood structure of nodes within a class.
\end{abstract}
\section{Introduction}
As GNNs are exploding in popularity and show widespread use cases ranging from recommender systems \mycite{wu_graph_2022} to predicting DNA-protein bindings \mycite{guo_dna-gcn_2021}, it has become increasingly important to understand how, and what, they learn. We consider the task of semi-supervised node classification (SSNC) using Graph Convolutional Neural Networks (GCN) \mycite{kipf_semi-supervised_2017} and aim to provide a holistic overview of how the GCN's learning of node embeddings and performance in SSNC depends heavily on underlying structures of the graph dataset itself. In particular, we will try to establish a link between a GCN's performance, node embeddings, and neighborhood structure of nodes within a class. To that end, we will use homophily as one of many metrics to analyse a graph's node neighborhood structure.
% This paper focuses on the exploring the importance of homophily within a graph structure of nodes in a semi-supervised node-classification task that infers the unknown labels of the nodes by using the network structure, given partially labeled networks with node features. Past works and research that explore the possibilities of better architectures in tasks of node-classification, seldom consider the impacts of nodes belonging the same class or have similar embedding features ie: homophily, on the effectiveness of graph machine-learning based tasks. We take semi-supervised node-classification as our graph based machine-learning task and analyse the impacts of node homophily ratios within the graph structures on the classification task. We use performance bench-marking as well as node-embedding cosine-similarity to support our investigation on how GCNs learn and their training and performance is impacted.
% \newpage
\section{Preliminaries}
\subsection{Homophily}
Homophily can be defined as the tendency of linked nodes to belong to the same class or have similar features in a graph-structure \mycite{zhu_beyond_2020}. For example, friends are likely to play similar sports and like similar genres of music. \newpage
We can define the node-homophily ratio as the following mathematical expression:
\begin{align*}
    \textbf{Node Homophily Ratio} = \frac{1}{|\mathcal{V}|} \sum_{v \in \mathcal{V}} \frac{|\{(w,v):w
    \in \mathcal{N}(v) \wedge y_v = y_w \}|}{|\mathcal{N}(v)|} \\\\
    = \frac{1}{|V|} \sum_{v \in V} \frac{\text { Number of } v^{\prime} s \text { neighbors who have the same label as } v}{\text { Number of } v ' s \text { neighbors }}
\end{align*}
\subsection{GCN}
We implement the GCN architecture \mycite{kim_how_2022} which can be described by the algorithm box provided below:
\begin{align*}
    Z=f(X)=\operatorname{softmax}\left(\operatorname{conv2}\left(\operatorname{ELU}\left(\operatorname{conv1}(X)\right)\right)\right)
\end{align*}
where $Z$ includes the classification of all the nodes with masked/ missing labels from the dataset $X$ (which contains node features and the edge list). We also add a dropout layer on the output of $\operatorname{ELU}$ before passing it on to $\operatorname{conv2}$.
\section{Related Works}
Papers have investigated the representation power of GCNs for SSNC \mycite{zhu_beyond_2020}. Building on previous literature claims, that many GNNs fail to generalize performance under heterophily, they propose a graph neural network, H2GCN that combines neighbor-embedding separation, higher-order neighborhoods, and intermediate representations to boost performance under heterophily. However, another paper \mycite{ma_is_2021} revisits the heterophilous graph-setting and finds that with some hyperparameter tuning, GCNs can outperform alternative methods uniquely designed to operate on heterophilous graphs by a sizeable margin. Besides showing that GCNs do not always “underperform” on heterophilous graphs, the paper also notes that in heterophilous graphs, if the neighborhood distribution of nodes with the same label (w.l.o.g. c) is approximately sampled from a fixed distribution $\mathcal{D}_{c}$ and different labels have distinguishable distributions, then GCN can excel at SSNC especially when node degrees are large. In our paper, we investigate how GCNs learn, and use the findings to find a plausible explanation for GCNs exhibiting such behaviour.
\section{Experiment}
We propose that for a given graph, the underlying distribution of a node's neighborhood has the most influence on the GCN's performance in SNCC.
\subsection{Dataset and setup}
We generate synthetic datasets, resembling the Cora dataset, using RandomPartitionGraphDataset class from Torch Geometric \mycite{hagberg_exploring_2008}. RandomPartitionGraph is a sampled synthetic graph of communities controlled by the node homophily and the average degree, and each community is considered as a class.
The node features are sampled from multivariate Gaussian distributions where the centers of clusters are vertices of a hypercube.
\\\\
We set the partitioning using the RandomSplit function, which performs a node-level random split and creates \verb|train_mask|, \verb|val_mask| and \verb|test_mask| in a $160 : 500 : 1000$ ratio.
\\\\
For our setup, we generate 9 synthetic datasets with homophily ratios ranging from 0.1 to 0.9. We fix 400 nodes per class and set average node degree to 10 while adopting other features from the real world Cora dataset (i.e. 7 classes and 1433 dimensional node features). We train separate GCNs with 128 hidden channels (i.e. the latent node embedding dimension) on the \verb|train_mask| for each generated dataset, to then test its performance in node classification on the \verb|val_mask| and \verb|test_mask|. \footnote{For the source code to reproduct experiments, please refer to \href{https://github.com/sannat17/CSC413-Project}{github.com/sannat17/CSC413-Project}}
\subsection{Results and Analysis}
\vspace{-10pt}
\begin{figure}[h]
\centering
\begin{subfigure}{.5\textwidth}
  \centering
  \includegraphics[width=1\linewidth,height=0.2\textheight]{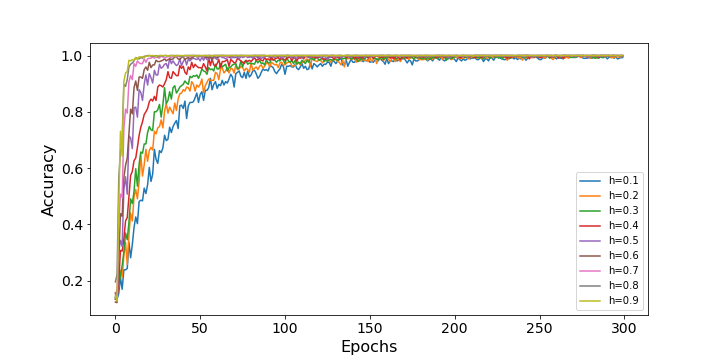}
  \caption{Training Accuracy}
\end{subfigure}%
\begin{subfigure}{.5\textwidth}
  \centering
  \includegraphics[width=1.0\linewidth,height=0.2\textheight]{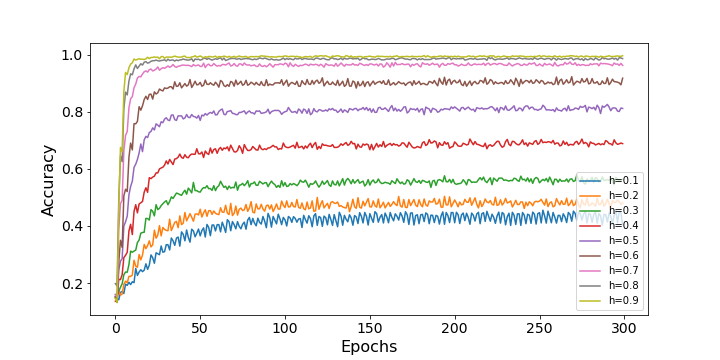}
  \caption{Validation Accuracy}
\end{subfigure}
\caption{\textit{Per epoch accuracy of GCNs trained on sythetic datasets with homophily ratio ranging from 0.1 to 0.9 for 300 epochs}}
\label{fig:test}
\end{figure}
In Fig. 1(b), we notice that a gradual increase in validation accuracy as the data set homophily increases. We are interested in understanding why does it become easier and faster for the GCN to learn the graph with high homophily coefficients. \\\\
To do so, we first, compare the node embeddings (NE) generated from the first convolutional layer for nodes with same and different classes (Fig 2). In order to compare these embeddings, we define the similarity score (C1NE) between vectors as the cosine of angle between them.\\
% ###############################################################################
\begin{figure}[h]
    \centering
    \includegraphics[width=0.5\linewidth,height=0.2\textheight]{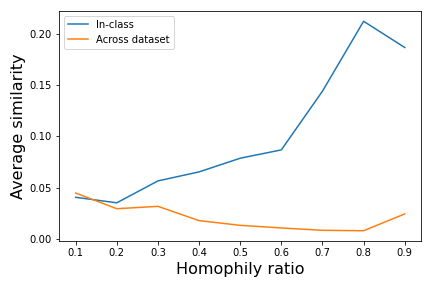}
    \caption{\textit{Average node-embedding cosine similarity for In-class and Across graph-datasets for node homophily ratios ranging from 0.1 to 0.9}}
    \label{fig:boat1}
\end{figure}\\
In order to analyse the GCN behaviour, WLOG, let $n_{1}, n_{2}$ be NE for two nodes with same label $x$ and $m$ to be the NE another node with label $y$ such that $x \neq y$. 
In Figure 2, we notice that when dataset homophily ratio = 1, C1NE$(n_{1}, n_{2}) \approx$ C1NE$(n_{1}, m) \approx$ C1NE$(m, n_{2})$. This implies, NE generated for nodes with same labels are as similar as NE for nodes with different labels. Thus, with very low similarity between NE for the same labels, the second convolution layer will not be able to learn classes effectively since there is no coherency to learn.
\\\\
Also, in Fig. 2, we can see as dataset homophily ratio increases, C1NE$(n_{1}, n_{2})$ increases and wlog, C1NE$(n_{1}, m)$ decreases. Moreover,  what is more interesting to analyse is the relative rapid increase in C1NE$(n_{1}, n_{2}) -$ C1NE$(n_{1}, m)$. That is, NE generated for nodes with same labels becomes more unique. Thus, progressively providing a more coherent embedding for NE of same class nodes and translating the second layer to learn more confident representations. 
\\\\
\textbf{Result 1:} We attribute these improvements in C1NE$(n_{1}, n_{2})$ to more consistent message passing which is caused by neighborhoods of same label nodes progressively being represented as samples of the same distribution and these neighborhood distributions becoming increasingly distinct for neighborhoods of nodes with different labels.
\\\\
We further break down GCN's learning by investigating it's performance across different node classes.
\begin{figure}[h]
\centering
\begin{subfigure}{.5\textwidth}
  \centering
  \includegraphics[width=1.2\linewidth,height=0.3\textheight]{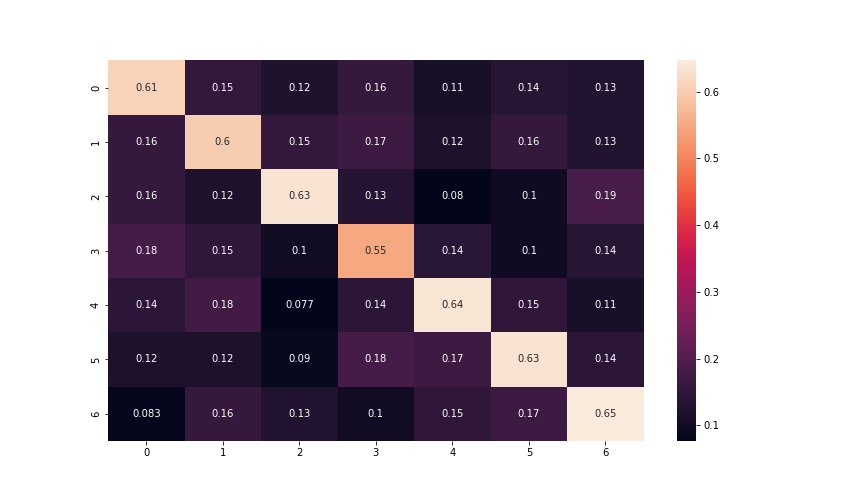}
  \caption{Homophily ratio 0.1}
\end{subfigure}%
\begin{subfigure}{.5\textwidth}
  \centering
  \includegraphics[width=1.2\linewidth,height=0.3\textheight]{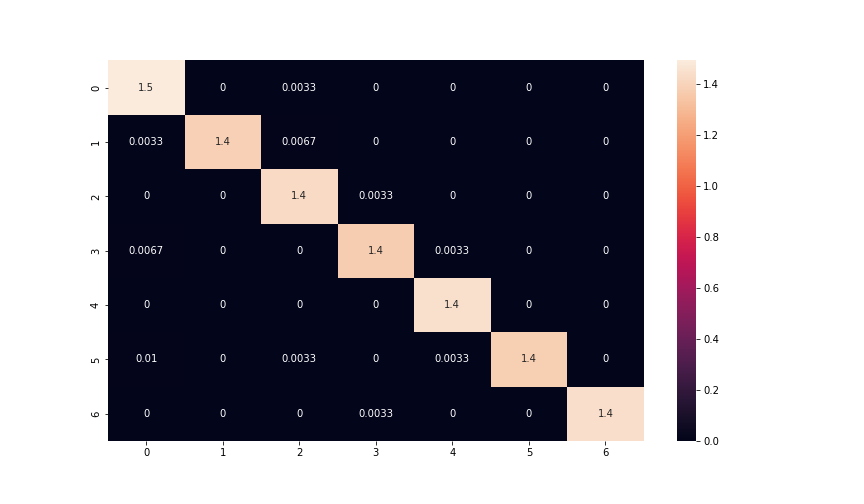}
  \caption{Homophily ratio 0.9}
\end{subfigure}
\caption{\textit{Confusion Matrix for different homophily ratio datasets}}
\label{fig:test}
\end{figure}
Figure 3 shows a significant improvement in confidence of SNCC classifications which reaffirms our previous conclusions about increasingly similar first layer node embeddings  translate into more confident learning. 
\\\\
\textbf{Result 2:} Another exciting observation is the uniformity of classification confidences; specifically we can conclude an invariable performance in classifying all 7 classes (observe how each predicted class in both confusion matrices have similar scores for correct classfications and equally distributed scores for incorrect classfications) from Fig 3. (a) and (b). We attribute both previously mentioned improvements in classification and confidences to the increase in neighborhood predictability.
\subsection{Conclusion}
Combining our findings, we conclude that while feature similarity for same label nodes is a foundational requirement for meaningful learning. GCN's SSNC performance is significantly influenced by consistency and uniqueness in neighborhood structure of nodes within a class (from Result 1 \& 2).
\\\\
These conclusions align with results of different papers \mycite{ma_is_2021}\mycite{zhu_beyond_2020} as high homophily in a graph is just a special case of unique and consistent neighborhoods. With our improved understanding, we can also claim that GCN's can perform well even under heterophily if the neighborhood distribution of nodes with the same label (w.l.o.g. c) is approximately sampled from a fixed distribution $\mathcal{D}_{c}$ and different labels have distinguishable distributions, as the first layer embedding would be similar for nodes with same labels.
\\\\
Lastly, we would like to note that to extend our work in understanding the learning of GCNs, future analysis can be done by varying average node degree, manipulating richness of node features, and using datasets with the special case of heterophily described above.
\\\\
\newpage
\printbibliography %Prints bibliography
%%%%%%%%%%%%%%%%%%%% Done %%%%%%%%%%%%%%%%%%%%%%%%%%%%%%%%%%
%%%%%%%%%%%%%%%%%%%%%%%%%%%%%%%%%%%%%%%%%%%%%%%%%%%%%%%%%%%%
\newpage
\appendix
\section{Appendix}
\subsection{Confusion Matrices}\vspace{-10pt}
\begin{figure}[h]
\centering
\begin{subfigure}{.5\textwidth}
  \centering
  \includegraphics[width=1.2\linewidth,height=0.3\textheight]{Training-Plots/confusion_0.1.png}
  \caption{Homophily ratio 0.1}
\end{subfigure}%
\begin{subfigure}{.5\textwidth}
  \centering
  \includegraphics[width=1.2\linewidth,height=0.3\textheight]{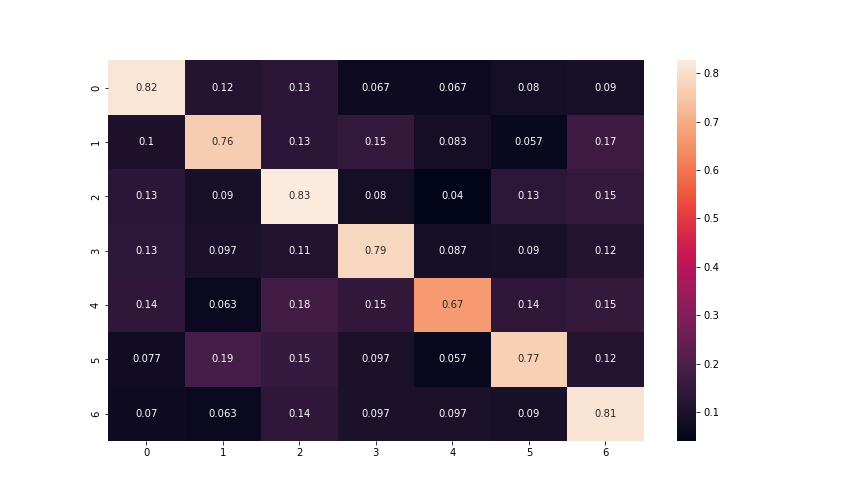}
  \caption{Homophily ratio 0.3}
\end{subfigure}
\begin{subfigure}{.5\textwidth}
  \centering
  \includegraphics[width=1.2\linewidth,height=0.3\textheight]{Training-Plots/confusion_0.3.png}
  \caption{Homophily ratio 0.7}
\end{subfigure}%
\begin{subfigure}{.5\textwidth}
  \centering
  \includegraphics[width=1.2\linewidth,height=0.3\textheight]{Training-Plots/confusion_0.9.png}
  \caption{Homophily ratio 0.9}
\end{subfigure}%
\caption{\textit{Confusion Matrix for different homophily ratio datasets}}
\label{fig:test}
\end{figure}
\newpage
\subsection{GCN Architecture}
\begin{figure}[h]
\includegraphics[width=0.9\linewidth,height=0.3\textheight]{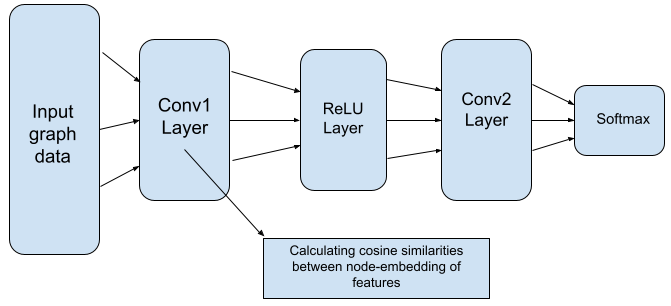}
\caption{GCN Architecture we used}
\label{fig:boat1}
\end{figure}
% Optionally include extra information (complete proofs, additional experiments and plots) in the appendix.
% This section will often be part of the supplemental material.
\end{document}